%% file: arxiv_paper.tex
\documentclass[runningheads]{llncs}
\usepackage{graphicx}
\usepackage{bbm}
\usepackage{comment}
\usepackage{bbold}
\usepackage{amsmath,amssymb} 
\usepackage{color, soul}
\usepackage{subcaption}

\usepackage{multirow}
\usepackage{titletoc}
\usepackage{graphicx}
\usepackage{multirow}
\usepackage[table,xcdraw]{xcolor}
\usepackage{booktabs}
\usepackage{arydshln}
\usepackage{breakcites}

\usepackage[thinc]{esdiff}
\usepackage{tablefootnote}
\usepackage{tabularx}
\usepackage{threeparttable}
\usepackage{footnote}
\usepackage{caption,pdfpages}
\usepackage{hyperref}
\usepackage[ruled]{algorithm2e}
\usepackage{minitoc}
\usepackage{blindtext}

\usepackage{listings}
\DeclareFixedFont{\ttb}{T1}{txtt}{bx}{n}{6} 
\DeclareFixedFont{\ttm}{T1}{txtt}{m}{n}{6}  

\definecolor{deepblue}{rgb}{0,0,0.5}
\definecolor{deepred}{rgb}{0.6,0,0}
\definecolor{deepgreen}{rgb}{0,0.5,0}

\definecolor{codegreen}{rgb}{0,0.6,0}
\definecolor{codegray}{rgb}{0.5,0.5,0.5}
\definecolor{codepurple}{rgb}{0.58,0,0.82}
\definecolor{backcolour}{rgb}{0.95,0.95,0.92}

\definecolor{codeblue}{rgb}{0.25,0.5,0.5}
\providecommand{\ie}[0]{\emph{i.e.}}
\providecommand{\eg}[0]{\emph{e.g.}}

        \makeatletter
        \renewcommand*\l@author[2]{}
        \renewcommand*\l@title[2]{}
        \makeatletter

\begin{document}
\pagestyle{headings}
\mainmatter
\def\ECCVSubNumber{929}  

\title{Inducing Predictive Uncertainty Estimation for Face Recognition}


\titlerunning{Inducing Predictive Uncertainty Estimation for Face Recognition}
%
\authorrunning{W. Xie \textit{et al.}}
\author{Weidi Xie$^1$, Jeffrey Byrne$^2$, Andrew Zisserman$^1$}
\institute{Visual Geometry Group, University of Oxford$^1$, Visym Labs$^2$\\[5pt]
\email{\{weidi,az\}@robots.ox.ac.uk, jeff@visym.com}}

\maketitle

\begin{abstract}
Knowing when an output can be trusted is critical for reliably using face recognition systems.
While there has been enormous effort in recent research on improving face verification performance, 
understanding when a model's predictions should or should not be trusted has received far less attention. 

Our goal is to assign a confidence score for a face image that reflects its quality in terms of recognizable information.
To this end, we propose a method for generating image quality training data
automatically from `mated-pairs' of face images, 
and use the generated data to train a lightweight Predictive Confidence Network,
termed as \emph{PCNet}, for estimating the confidence score of a face image.
We systematically evaluate the usefulness of PCNet with its error versus reject performance, 
and demonstrate that it can be universally paired with and improve the robustness of any verification model.  
We describe three use cases on the public IJB-C face verification benchmark: 
(i) to  improve 1:1 image-based verification error rates by rejecting low-quality face images; 
(ii) to improve quality score based fusion performance on the 1:1 set-based verification benchmark; 
and (iii) its use as a quality measure for selecting high quality (unblurred, good lighting, more frontal) faces from a collection,
\eg~for automatic enrolment or display.

\end{abstract}

\section{Introduction}
\label{sec:intro}
\input{text/introduction.tex}

\section{Related Work}
\input{text/related_works.tex}

\section{Approach}
\label{sec:method}
\input{text/method.tex}

\section{Experiments}
\label{sec:exp}
\input{text/experiments.tex}
\input{text/results.tex}


\section{Conclusions}
\input{text/conclusions.tex}

\section*{Acknowledgment}
This research is based upon work supported by the Office of the Director of National Intelligence (ODNI), Intelligence Advanced Research Projects Activity (IARPA), via contract number 2014-14071600010. 
The views and conclusions contained herein are those of the authors and should not be interpreted as necessarily representing the official policies or endorsements, either expressed or implied, of ODNI, IARPA, or the U.S.\  Government.  
The U.S.\  Government is authorized to reproduce and distribute reprints for Governmental purpose notwithstanding any copyright annotation thereon. 
Funding for this research is also provided by the EPSRC Programme Grant Seebibyte EP/M013774/1. 

\clearpage

\bibliographystyle{splncs04}
\bibliography{shortstrings,vgg_local,vgg_other}

\clearpage
\appendix
\noindent {\Large\textbf{Appendix}}

\section{Architecture for PCNet}
Here, we present the backbone architecture for the proposed PCNet,
which is based on the standard ResNet18 with an extra fully connected layer at the end.
\input{tables/architecture.tex}

\clearpage
\section{More visualization of single-image confidence scores}

\begin{figure*}[!htb]
\footnotesize
\begin{center}
\includegraphics[width=\textwidth]{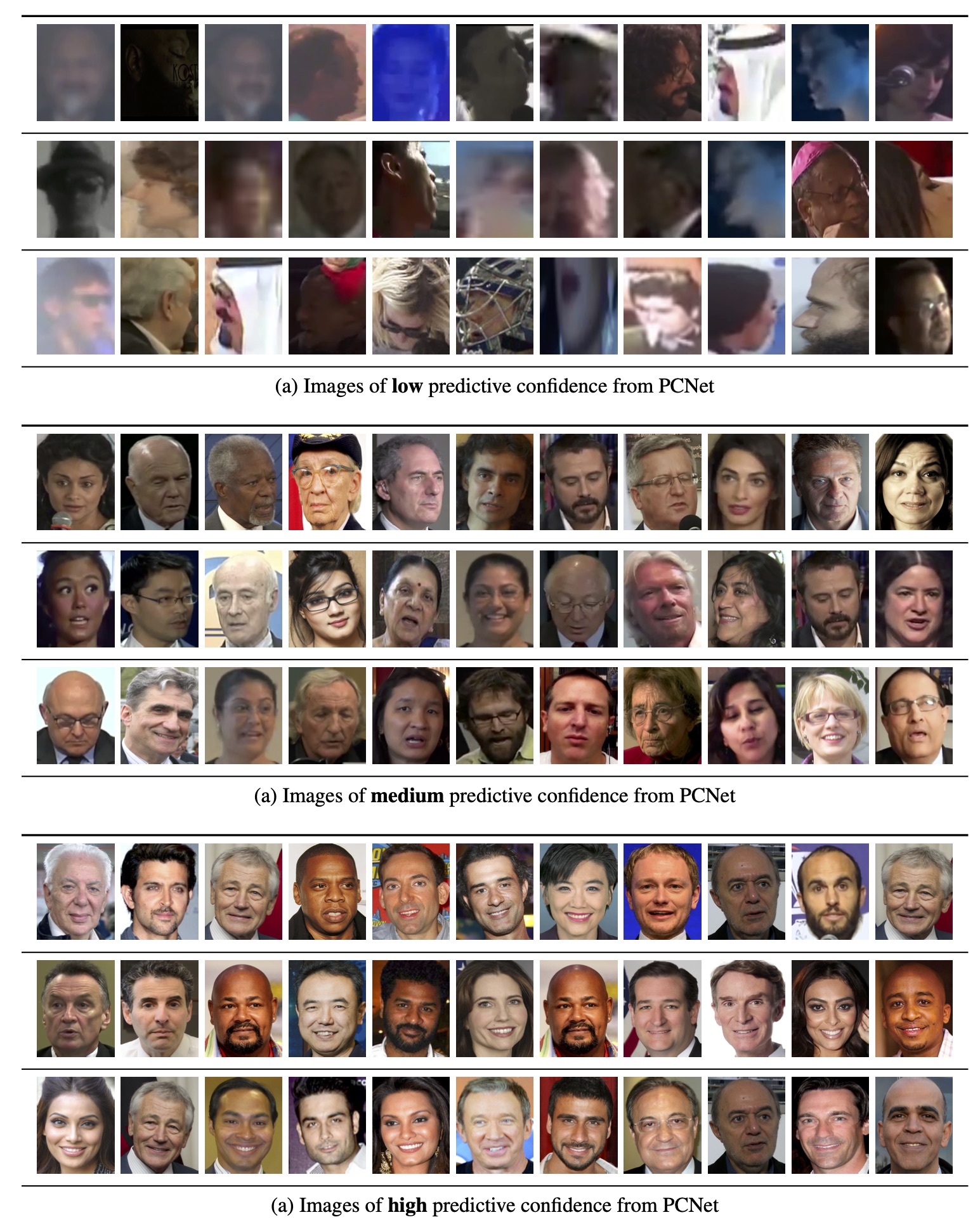}
\vspace{-10pt}
\caption[]{After ranking all images of the IJB-C datasets, 
we split the ranking into three different ranges, and randomly sample images from the corresponding range.}
\vspace{-.5cm}
\label{fig:vis}
\end{center}
\end{figure*}

\clearpage
\section{Visualizing the rejected pairs by PCNet}
\begin{figure*}[!htb]
\footnotesize
\begin{center}
\includegraphics[width=\textwidth]{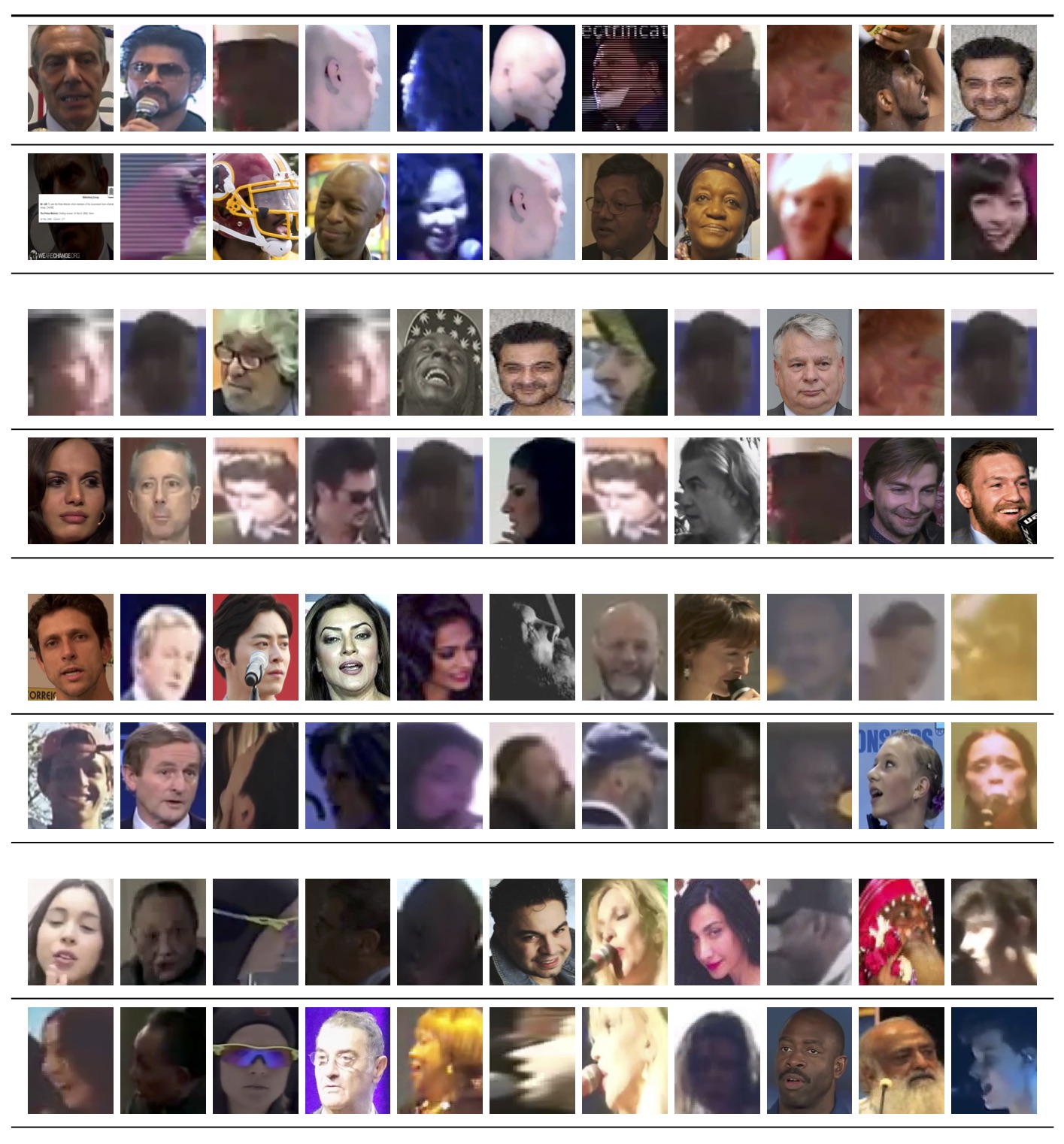}
\vspace{-10pt}
\caption[]{Visualizing the rejected pairs by PCNet while ploting the error vs rejection curve. 
Note that, the figures are grouped by pairs in the column, \ie~two images are shown in each pairs.}
\vspace{-.5cm}
\label{fig:vis2}
\end{center}
\end{figure*}

In Figure~\ref{fig:vis2},
we show example pairs that have been rejected while ploting the error vs rejection curve,
in other words, 
these pairs are predicted with the lowest predictive confidence from our PCNet.
As expected, 
once single image or both images are of low quality in the sense of insufficient information to be recognizable, 
the PCNet can indeed alarm the users.

\end{document}

%% file: text/introduction.tex

There has been tremendous progress in face recognition over the past five years, 
primarily due to three factors:
\emph{First}, 
the development of neural network architectures, from AlexNet~\cite{Krizhevsky12}, 
to VGGNet~\cite{Simonyan15}, to ResNet~\cite{He16};
\emph{Second},
the introduction of  more sophiscticated objective functions, 
for instance, contrastive loss~\cite{Chopra05}, triplet loss~\cite{Weinberger05}, large-margin softmax~\cite{Liu17,Deng19}. 
\emph{Third},
the large-scale datasets, \eg~VGGFace~\cite{Parkhi15}, UMDFace~\cite{Bansal16}, MS1M~\cite{Guo16}, 
VGGFace2~\cite{Cao18}, IMDB-Face~\cite{Wang18}, 
that have enabled the data-hungry neural network models to be trained.
With these efforts, 
state-of-the-art face recognition models have  demonstrated strong capabilities of learning effective identity embeddings,
which are largely invariant to nuance factors, such as pose and age, yet are still discriminative for face identities.

\input{figures/teaser.tex}

In this paper our objective is face {\em verification} -- the task of determining if two face images are of the same
person or not; or, more generally, given two sets of faces, 
where each set only contains arbitrary number of images from one person, 
determine if the two sets are of the same person or not.
Verification, as is this case for any face recognition task, depends on the assumption that 
the input image  contains sufficient  information to be recognizable.
During the training stage, 
this assumption is usually guaranteed, largely due to the bias from the data collection process --
that  images have been curated by human annotators, and so must be recognizable.
Unfortunately, this is not always the case during the inference stage, 
as the input face  images to verification system may be in profile, blurry, or low resolution
(or even non-face images if the face detector operating point is for very high recall).
As a result, 
this train-test discrepancy will potentially lead to false positives or negatives during verification.
For instance, 
as demonstrated in Figure~\ref{fig:teaser},
a well-trained face recognition model~(ResNet101 in this case) is broken by one low-quality image.
Conceptually, this challenge can be resolved by augmenting the similarity between the
face embeddings with a {\em predictive confidence},
which is identity-agnostic, 
and only reflects whether the image contains sufficient discriminative information to be recognizable.

Estimating such confidence scores is a non trivial task, 
as it is costly and challenging to obtain groundtruth annotations.
Indeed, even defining image quality  is difficult,
despite the early efforts~\cite{ISO/IEC11,ICAO15} on measuring image quality by pose, 
expression, illumination, occlusion, and face accessories, 
some metrics remain extremely subjective. 
Furthermore, 
since classifications can be changed by adding perturbations to images (adversarial attacks) that 
are indistinguishable to human observers~\cite{Szegedy14,Goodfellow15,Anh15,Rozsa16}, 
human assessments of image quality may be only sub-optimal for network training.

In this paper, 
we propose a method for generating image quality training data automatically, and use
the generated data to train a lightweight network  to predict confidences for any face image.
The only requirement of the method is to have sets of images of the same person -- and such sets are
readily available from public face datasets that have identity annotation. 
Once the Predictive Confidence Network, {\em PCNet}, 
has been trained, then it can be applied to any verification system and any face images.
This method is described in Section~\ref{sec:method}.
In Section~\ref{sec:exp}, 
we systematically evaluate the usefulness of  PCNet with error versus reject curves~\cite{Grother07}.
Experimentally, 
we demonstrate it can be universally paired with and improve the robustness of other recognition models, 
including strong models such as SENet50 and ResNet101, 
and that  PCNet outperforms previous quality estimation baselines, 
while using a significantly lighter architecture~(ResNet18 vs ResNet50).
We also demonstrate three use cases on the challenging JANUS IJB-C Benchmark~\cite{Maze18}, 
(i) PCNet can be used to significantly improve 1:1 image-based verification error rates, 
such as False Accept Rate~(FAR), and True Accept Rate~(TAR), 
of automatic face recognition systems by rejecting low-quality face images;
(ii) it can be used for quality score based fusion where a weighted average is used to combine 
the descriptors of multiple images of the same face into a single descriptor,
significantly improve the performance on the 1:1 set-based verification benchmark; 
and (iii) it can also be used as a quality measure for selecting good (unblurred, good lighting, more frontal) faces from a collection, 
\eg~for automatic enrollment or display.

%% file: figures/teaser.tex
\begin{figure*}[t]
\footnotesize
\begin{center}
\begin{tabular}{cccc}
\hspace{0.5cm}
\includegraphics[width=0.14\textwidth, height=0.18\textwidth]{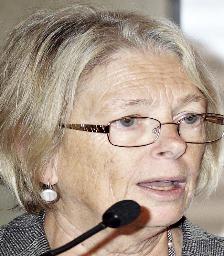} & 
\includegraphics[width=0.14\textwidth, height=0.18\textwidth]{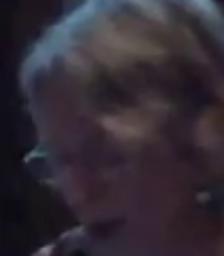} &
\hspace{1.5cm}
\includegraphics[width=0.14\textwidth, height=0.18\textwidth]{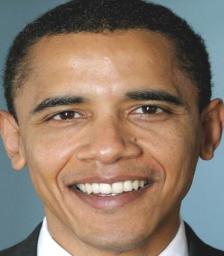} & 
\includegraphics[width=0.14\textwidth, height=0.18\textwidth]{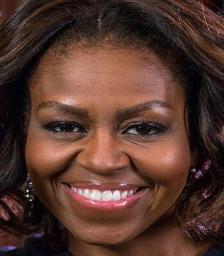} \\[2pt]
\multicolumn{2}{c}{\footnotesize{(a) Mated pair~(similarity score$\approx$0.4)}} &
\multicolumn{2}{c}{\footnotesize{\hspace{0.5cm}(b) Non-mated pair~(similarity score$\approx$0.4)}}\\[2pt]
\multicolumn{2}{c}{\footnotesize{\underline{Predictive Confidence = 0.13}}} &
\multicolumn{2}{c}{\footnotesize{\hspace{0.5cm}\underline{Predictive Confidence = 0.95}}}\\
\end{tabular}
\caption[]{A face verification model~(ResNet101 trained on VGGFace2) gives similar output scores to both pairs, 
In (a), it refers to a false negative matching, 
where the low similarity score is most likely due to the inadequate information in the second image.
In (b), the similarity score indicates that the identities are different in the pair of images.
Predictive confidence is required to decide whether to trust the output from the system or not.}
\label{fig:teaser}
\vspace{-0.7cm}
\end{center}
\end{figure*}

%% file: text/related_works.tex
\vspace{-0.2cm}
\paragraph{Learning set representation.}
Recent works have proposed architectures for learning face descriptor aggregation~\cite{Yang2017,Tran2017,Xie18a,Xie18b}.
The general idea is to compute a set representation by the weighted average of the individual face,
where the weights are treated as a latent variable inferred from the deep networks.
While optimizing for classification,
the training process implicitly tries to suppress the contribution from low-quality images, 
and highlight the most discriminative face images.
This has later been interpreted as fulfiling the function of quality estimation. 
Despite the results shown in~\cite{Xie18b}, 
these methods lead to over-confident predictions,
\ie~majority of the images will have a high quality score.

\vspace{-0.3cm}
\paragraph{Learning with rejection.}
In the cases of learning with single instance,
the problem of classification with a reject option or learning with abstention~\cite{Grandvalet09,Yuan10,Cortes16} is highly related, where the classifier is allowed to abstain from making a prediction at a certain cost. 
Typically such methods jointly learn the classifier and the rejection function. 
Our paper aims to provide a standalone model that enables to learn the confidence scores independently to 
any already trained and possibly black-box face recognition systems. 
Technically,  our goal is to learn an appropriate ranking for the confidence scores for the images,
but we do not explicitly learn the appropriate rejection thresholds.

\vspace{-0.3cm}
\paragraph{Visual quality estimation.}
In biometric recognition, 
image or sample quality has long stood out as the obvious way of predicting system performance~\cite{Li05a,Scheirer11}, 
where poor-quality images pose significant challenges.
Traditionally, 
quality estimation has focused on the image capturing requirements defined by humans, 
for instance, in ISO/IEC 19794-5~\cite{ISO/IEC11},  ICAO 9303~\cite{ICAO15},
the quality is usually measured by pose, expression, illumination, occlusion, and accessories.
In the recent literature, learning-based approaches start getting popular, \eg~\cite{Abaza12,Abaza14,Abhishek14,Hsu06,Phillips13,Aggarwal11,Wong11,Chen15,Kim15,Best-Rowden18,Hernandez-Ortega19,Terhorst20}.
See~\cite{Best-Rowden18,Terhorst20} for an excellent extended literature review.


%% file: text/method.tex
In this section we describe the \textbf{Predictive Confidence Network}~(PCNet), that ingests a face image
and outputs a scalar indicating the likelihood of the face being identifiable by a state-of-the-art face verification system.
The training method proceeds in two stages: first, there is a 
simple and scalable approach for generating {\em pairwise} verification scores  using only 
mated face-image pairs, \ie~face images of the same person. 
Second, we provide an approach to disentangle  the pairwise scores to enable training of the PCNet for single faces.
We illustrate the method using the VGGFace2 dataset (described in Section~\ref{sec:datasets}) 
which is partitioned here into two halves by identites,  with the faces of around 4300 identities in each part.

\subsection{Generate Pairwise Verification Scores}
\label{sec:gen}
A standard ResNet34 is trained for face classification on the first half of the dataset, \ie~a 4300-way classification.
Once trained, verification scores are obtained for all mated pairs in the the other half of the dataset, as shown in Figure~\ref{fig:generate_pc}.
We also alternate this process (\ie~training the ResNet34 on the other half, etc)
in this way we obtain pairwise verification scores for all mated pairs in the dataset, 
ending up with roughly 500 million pairwise scores in total. 
The scores are obtained in this way, using the two
halfs of the data, so that the scores are not obtained from the same samples that the network is trained on.
Note, the verification score here is obtained as the cosine similarity between the face embeddings, 
with a score of $1.0$ indicating a perfect match.
\input{figures/pipeline.tex}

\subsection{Training PCNet}
\label{sec:pcnet}
We make the assumption that if the verification score is less that $1.0$, 
then this is due to recognizability information being missing from either of the images forming the pair.  
We then use a `loser takes all' scheme to obtain the quality measure for the individual images of the pair.
That is to say, 
we assume the pairwise verification score is fully determined by the image with worst quality (or least
discriminative information).
This then becomes a training target for the PCNet: 
it is trained to output a predictive confidence for each image of the pair, 
such that the minimum of the two confidences equals the verification score of the pair.

Formally, during training, 
a mated pair of images is selected (\ie~both images are of the same identity)
and each image is passed through PCNet, 
parametrized as~$\Phi(\cdot)$, and outputs a scalar~$s$, 
referred as the predictive confidence for the image.
For the two images of the pair, if 
$s_1 = \Phi(I_1; \theta)$ and $s_2 =  \Phi(I_2; \theta)$, then the 
training objective for optimization is defined as the mean square loss,
where $y$ is the verification score of the image pair, 
mathematically, we minimize the following loss:
\[
\mathcal{L}(s_1,s_2) = \mathbbm{1}\{s_1 < s_2\} \cdot |s_1 - y|^2 + \mathbbm{1}\{s_2 < s_1\} \cdot |s_2 - y|^2
\]
where $\mathbbm{1}\{\cdot\}$ refers to the indicator function. 
Note that this loss guarantees the permutation invariance between the images in the mated pair.
The PCNet is then trained with over 500 million pairwise scores.
For simplicity, in this paper, a standard ResNet18 is used as $\Phi(\cdot)$,
but the proposed method is not limited to any specific architecture.


%


%% file: figures/pipeline.tex
\begin{figure*}[!htb]
\footnotesize
\begin{center}
\begin{tabular}{ll}
\hspace{-.3cm}
\includegraphics[height=0.28\textwidth]{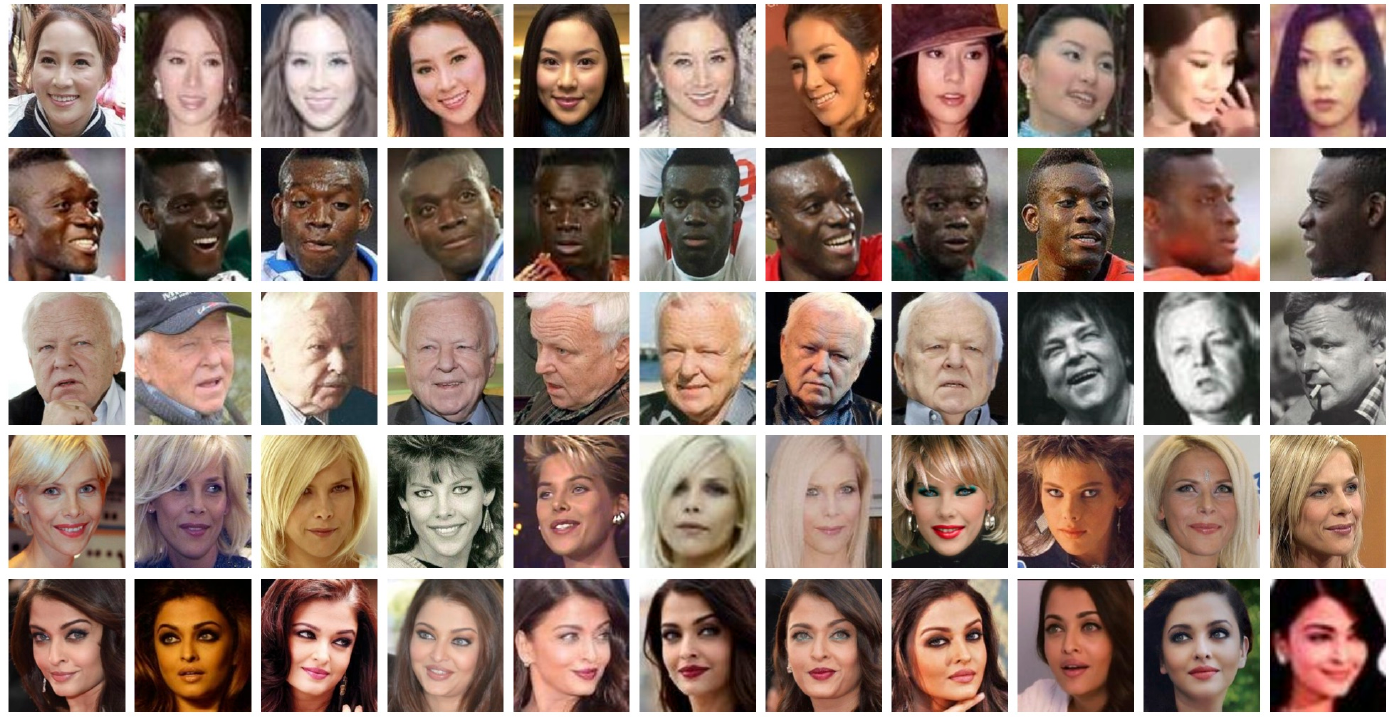} &
\hspace{.3cm}
\includegraphics[height=0.29\textwidth]{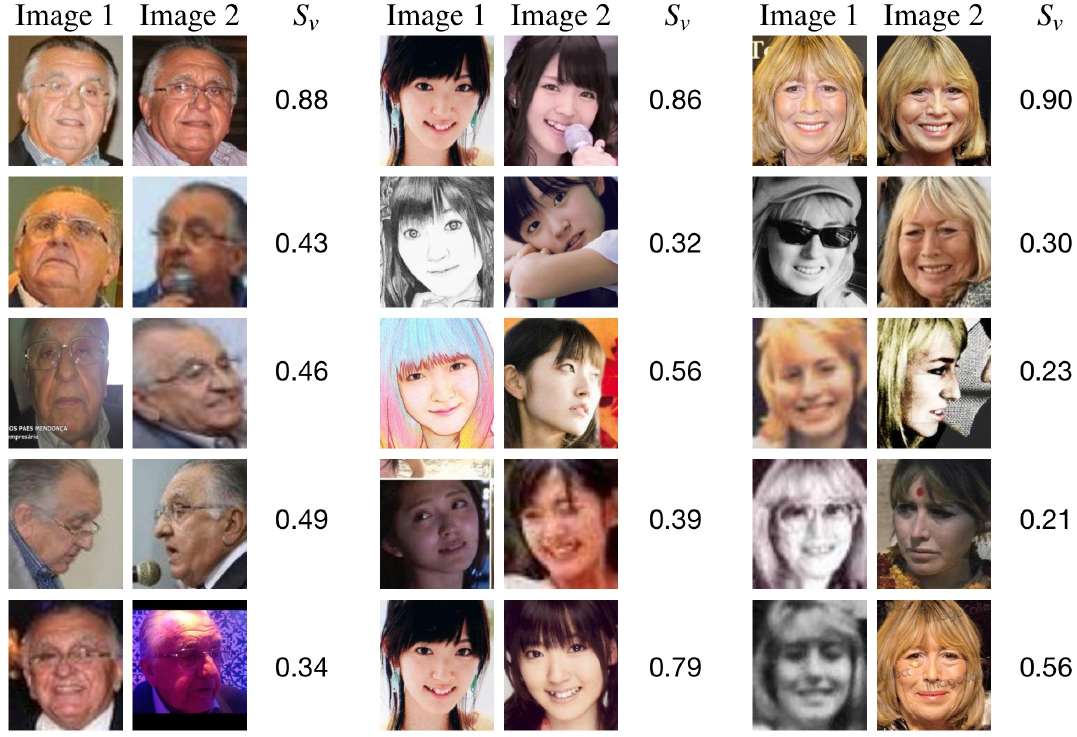} \\
\multicolumn{1}{c}{\footnotesize{(a) Train ResNet34 for classification.}} &  
\multicolumn{1}{c}{\hspace{.1cm}\footnotesize{(b) Verification scores for mated pairs.}}
\end{tabular}
\vspace{-5pt}
\caption[]{
\footnotesize{Generating verification scores for mated pairs.}}
\label{fig:generate_pc}
\end{center}
\vspace{-1cm}
\end{figure*}

%% file: text/experiments.tex
\subsection{Datasets}
\label{sec:datasets}
\paragraph{VGGFace2~\cite{Cao18}} is used through this paper, to train all face recognition models and PCNet.
It contains about $3.31$ million images with large variations in 
pose, age, illumination, ethnicity and profession (e.g.~actors, athletes, politicians). 
Approximately, $362.6$ images exist for each of the $9131$ identities on average.
In order to be comparable with existing models, 
we follow the same dataset split and only train on the training set~($8631$ identities).

\paragraph{IJB-C Dataset~\cite{Maze18}}  is used for all the evaluations in this paper,
it is a superset of the previous IJB-A and IJB-B datasets.
Overall, it contains $3,531$ subjects with $31.3$K still images and $117.5$K frames from $11,779$ videos,
captured from unconstrained environments with large variations in viewpoints, image quality and distractors~(non-face images).
It is generally considered as one of the most challenging \emph{public} benchmarks for face recognition.

\subsection {Training Details} 
While generating pairwise predictive confidence and training PCNet,
we follow the same strategy, namely, 
resizing the shorter side to $256$,
and a region of $224\times 224$ pixels is randomly cropped from each sample.
The mean value of each channel is subtracted.
Stochastic gradient descent is used with mini-batches of size $256$, 
with a balancing-sampling strategy for each mini-batch due to the unbalanced training distributions.
The initial learning rate is $0.1$ for the models learned from scratch, 
and this is decreased twice with a factor of $10$ when errors plateau. 
As for augmentation during training ResNet34~(in Section~\ref{sec:gen}),
random transformations are used with a probability of $20\%$ for each image,
e.g.\ monochrome augmentation, horizontal flipping, and geometric transformation.
As for generating the pairwise predictive confidence pseudo-groundtruth, 
each image in the mated pair can potentially have a probability of $0.2$ of
randomly picking at least one of the degradations from Gaussian blur, motion blur, and jpeg compression,
and the degraded images are later used for training PCNet.

\subsection{Evaluation Protocol}
We benchmark on 1:1 covariate verification and 1:1 verification from JANUS IJB-C.
The former refers to the popular image-to-image verification, 
while the latter refers to  set-to-set verification, 
where each set could potentially contain any number of images of the same identity.
For both cases, 
the performance is reported as the standard True Accept Rate~(TAR) vs.\ False Accept Rate~(FAR)
(\ie receiver operating characteristics (ROC) curve). 

To evaluate the effectiveness of `predictive confidence', 
we report the error versus reject curves for 1:1 covariate verification~(Section~\ref{sec:error_vs_reject}),
a metric originally proposed for measuring biometric quality~\cite{Grother07},
and recently adopted for face recognition~\cite{Grother19}.
These curves show a verification error-rate over the fraction of ignored face images. 
Based on the predictive confidences values, 
these rejected images are those with the lowest confidences and the error rate is calculated on the remaining images. 
The curves indicate good quality estimation
when the verification performance increases (the error decreases) monotonically as more images are rejected (as this indicates that the uninformative images are being rejected first).
This process allows a fair comparison to different algorithms for face quality assessment, 
since it is independent of the range of the quality predictions (only the ordering is used).

As PCNet only provides predictive confidence, but not verification functionality,
it is coupled with three different open-source face recognition models that are publicly available,
namely, ResNet50, SENet50 and ResNet101, for verification.
Although these models have all been trained on VGGFace2,
they do behave slightly differently  as the training settings vary, \eg~data augmentation, learning rate schedule, \emph{etc}, as well as due to the differences in the architectures.
Note that, the purpose of this paper is not to benchmark the state-of-the-art face recognition models, 
instead, we aim to validate the conjecture that the predictive confidence is an effective component for different models, 
\ie~largely model-independent.

\subsection{Baselines}
We compare with two recent works~\cite{Xie18b,Hernandez-Ortega19}, 
which propose the idea of using image quality estimation to improve face recognition systems.
In~\cite{Xie18b}, the authors propose the Multicolumn Networks~(\emph{MNet}), 
learning a set representation through a weighted average of all individual images in the set.
As a by-product, the networks learn  a quality estimation that pays more attention to images with more discriminative information, \eg~frontal faces, high-resolution images.
In~\cite{Hernandez-Ortega19},
Face-QNet~(\emph{QNet}) is trained by comparing images with some `golden' reference images, 
that were selected by  ICAO Compliance Software. 
We use the official implementation and models~\cite{qnet}.
In both works
the quality estimation models were trained on VGGFace2 dataset with a ResNet50 architecture;
however, in the following sections,
we demonstrate that our light-weight PCNet~(based on ResNet18) outperforms these strong baseline models on all metrics.

%% file: text/results.tex
\subsection{Results: 1:1 Covariate Verification with Rejection}
\label{sec:error_vs_reject}
In this protocol, 
the goal is to perform still image-to-image verification.
In total, there are $140$K images with over $7$M genuine matches, and $39$M impostor matches.
During inference, 
we define the predictive confidence for each pair of images as the minimum score of the two images,
and rank these pairwise confidence scores in descending order.
By rejecting the bottom $k\%$ pairs, where $k \in [0,  40]$,
the obtained TAR and FAR will therefore be informative to understand 
if the failure cases from modern face recognition systems are indeed predicable from the confidence scores.

\input{tables/IJBC_11_covariate.tex}

\begin{figure*}[!htb]
\footnotesize
\begin{center}
\includegraphics[width=\textwidth]{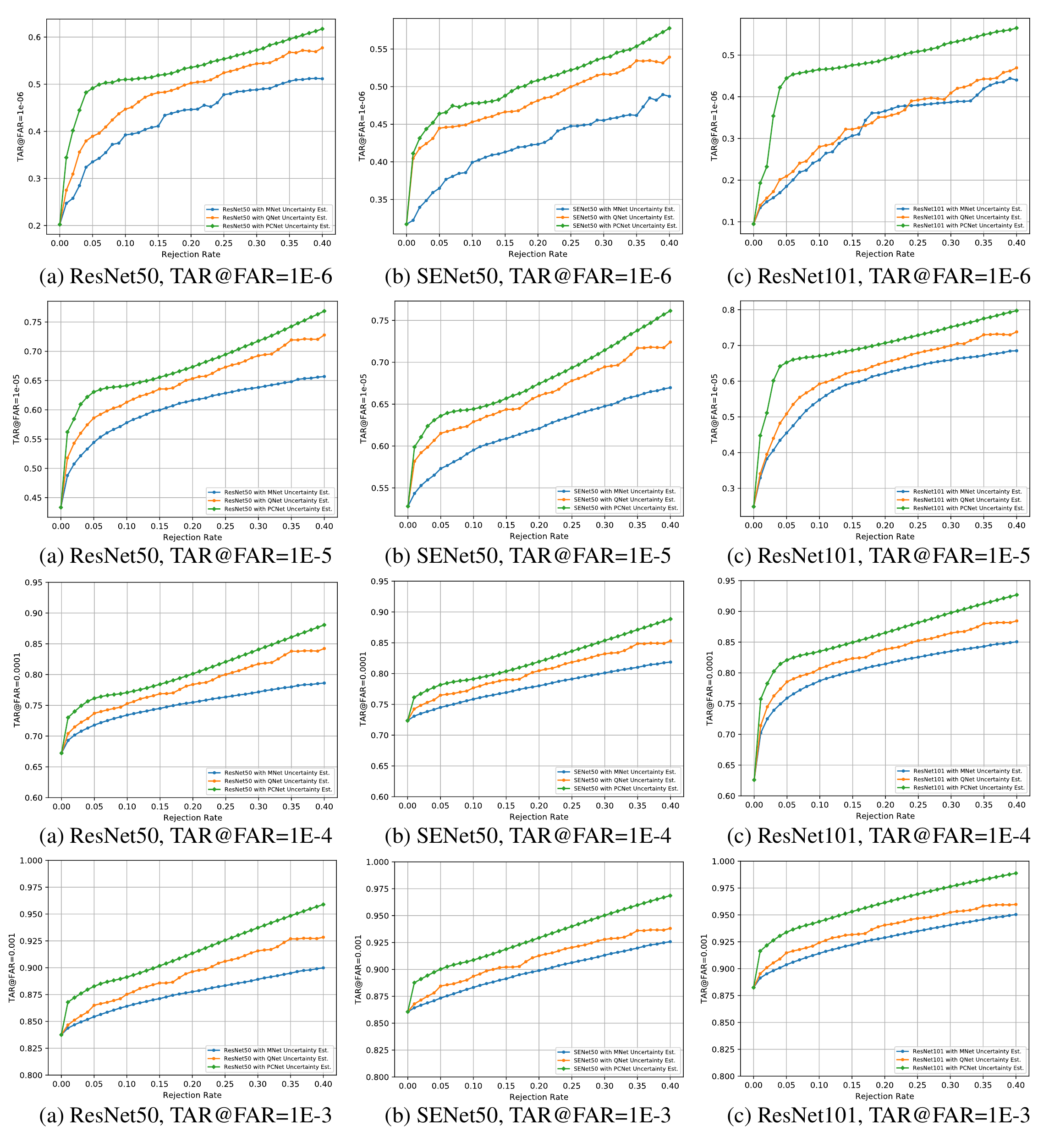} 
\vspace{-20pt}
\caption[]{
\footnotesize{Error vs rejection curve on IJB-C 1:1 covariate verification, 
with rejection rate being densely varied.}}
\vspace{-20pt}
\label{fig:ijbc-11cov1}
\end{center}
\end{figure*}

As shown in Table~\ref{tab:11cov}, 
we sample five different rejection rates~($[0.0, 0.4]$),
where $0.0$ refers to the case where no pairs are rejected, 
\ie~the performance from raw verfication systems~(ResNet50, SENet50 and ResNet101).
More complete results are shown in Figure~\ref{fig:ijbc-11cov1}, 
where rejection rates are densely sampled with a gap of $0.01$. 
It is clear that
when rejecting the face pairs with lowest predictive confidences, 
the performance of all face recognition systems has been improved significantly.
This claim holds for all the different architectures, 
demonstrating the generalizability of the PCNet --
meaning it is largely recognition  model independent.
When comparing with baseline models~(MNet and QNet), the proposed PCNet shows superior performance on all metrics.

\begin{figure*}[!htb]
\footnotesize
\begin{center}
\hspace{0.0cm}
\includegraphics[width=\textwidth]{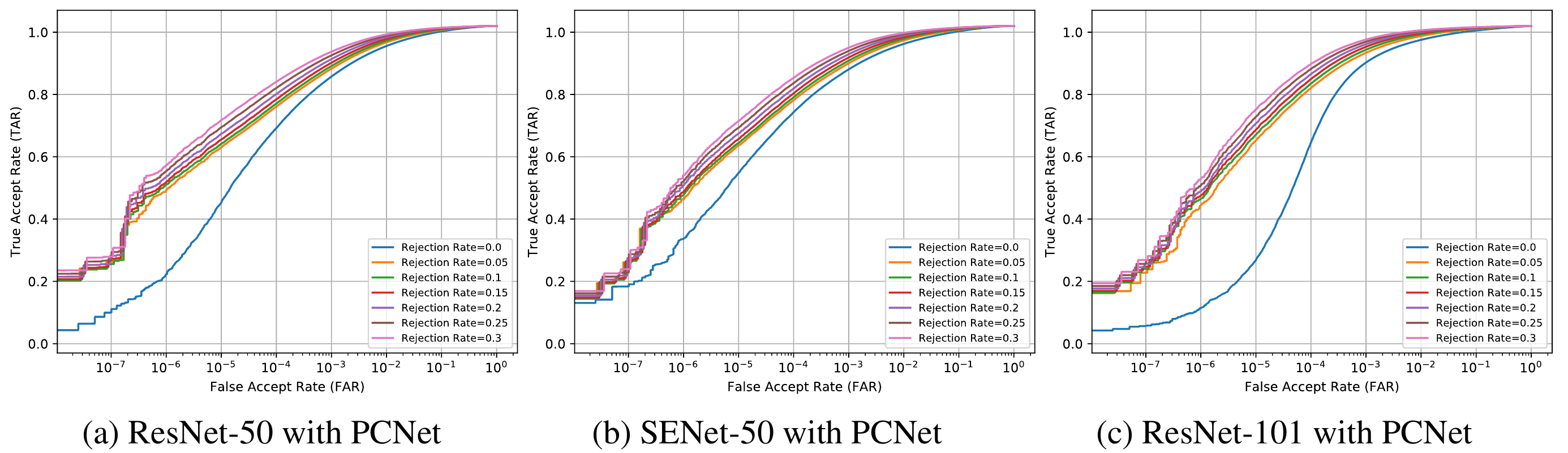} 
\vspace{-15pt}
\caption[]{
\footnotesize{
ROC curves for IJB-C 1:1 covariate verification. 
Benchmarked for different architectures under different rejection rate.
As can be seen, 
while only rejecting $5\%$ of the pairs with lowest predictive confidences,
PCNet can already improve the verification performance significantly.}}
\label{fig:ijbc-11cov2}
\vspace{-.8cm}
\end{center}
\end{figure*}

In Figure~\ref{fig:ijbc-11cov2},
we plot the complete ROC curve on 1:1 covariate verification.
It is interesting to see that
PCNet is already very effective when only rejecting $5\%$ of the pairs with lowest predictive confidences.
Remarkably, the verification performance of ResNet101 has been boosted around $20\%$ for TAR@FAR=1E-6 and TAR@FAR=1E-5, 
suggesting that PCNet is indeed producing informative confidence scores 
that reflect the potential limitations of modern face reconigiton systems.

\begin{figure*}[!htb]
\footnotesize
\vspace{-.5cm}
\begin{center}
\includegraphics[width=\textwidth]{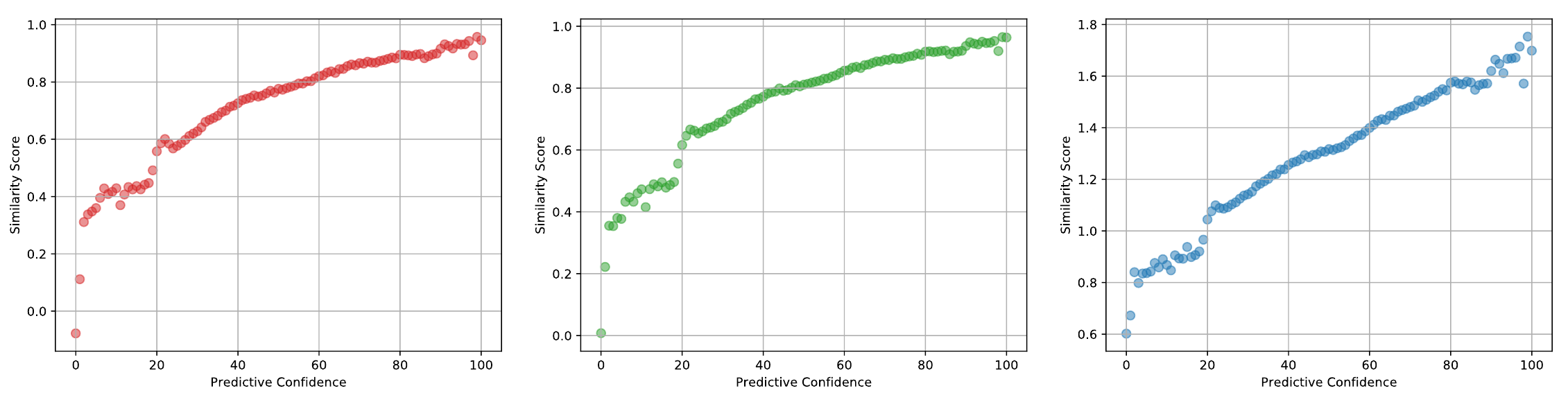} 
\vspace{-15pt}
\caption[]{
\footnotesize{The correlation between predictive confidence and the similarity scores from different models. 
The predictive confidences correlate with the similarity scores from different face verification systems -- even though it was not trained on them. }}
\vspace{-1cm}
\label{fig:ijbc-11cov3}
\end{center}
\end{figure*}

In Figure~\ref{fig:ijbc-11cov3},
we plot the correlation between the predictive confidence and the similarity scores from different models,
we split the confidence scores into $100$ bins, and compute the mean similarity scores falling in each bin.
Note that the similarity scores are only generated for the mated pairs in the IJB-C,
as such curve for non-mated scores will not be informative,
because it is expected that all points lying on a narrow band on the very left side,
either due to low predictive confidence or high predictive confidence by different identities.
From the strong correlations between predictive confidence and matching scores for all models, 
it shows that the proposed PCNet is effective for avoiding false negatives during evaluation, 
and also it is model independent, shown from the consistent correlation among different models.

\paragraph{Discussion.} 
These results demonstrate that PCNet has indeed learnt a quality measure that correlates
with the information content of the face image. 
Note in particular, 
the error vs.\ reject curves have exhibited a monotonic improvement with increasing rejection ratio,
meaning that the rejected pairs are of the low visual quality that face recognition systems struggle on.
Despite the fact that the PCNet is only trained with mated pairs, 
verification evaluation suggests that it also successfully orders the non-mated pairs.
Examples of the rejected pairs~(including mated and  non-mated) will be given in the arXiv version.
 
\subsection{Results: Standard 1:1 Verification with Confidence Weighting}
This protocol uses set-to-set verification, 
where each set consists of a variable number of face images and video frames from different sources:
each set can be image-only, video-frame-only, or a mixture of still images and frames.
This protocol defines $23124$ different sets, with $19557$ genuine matches, and over $15$M impostor matches.
During testing, the set descriptor is computed as a weighted average of individual faces,
with the weights obtained from the  predictive confidences of the faces as 
$v = \sum_i s_i \cdot v_i / \sum_i s_i$
where $s_i, v_i$ refers to the predictive confidence and feature embedding for the image~($I_i$).

\begin{figure*}[!htb]
\footnotesize
\begin{center}
\includegraphics[width=.9\textwidth]{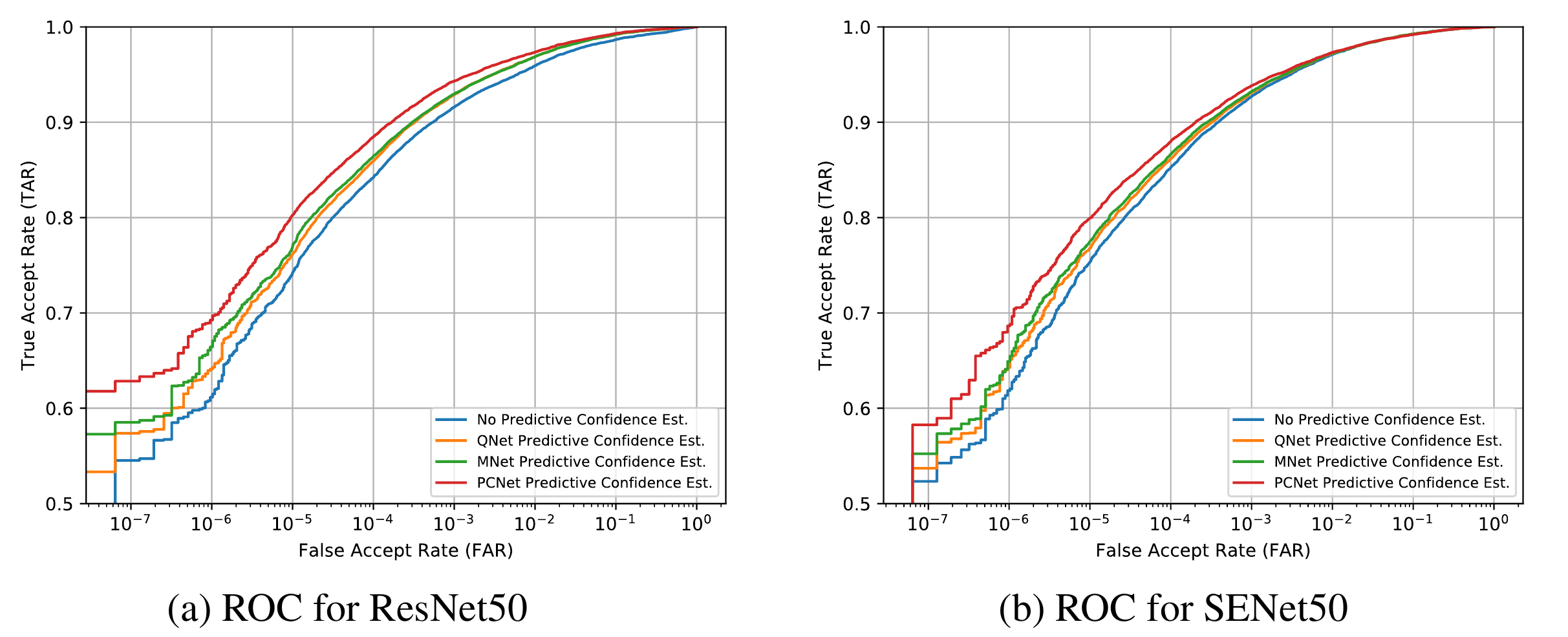}
\vspace{-6pt}
\caption[]{
\footnotesize{On 1:1 IJB-C Verification,
the PCNet improves the set-based verificationc for different face verification architectures, 
and outperform both QNet and MNet.}}
\label{fig:ijb-roc}
\vspace{-10pt}
\end{center}
\end{figure*}

\paragraph{Discussion.}
As shown in Table~\ref{tab:ijbc-evaluation-table2} and Figure~\ref{fig:ijb-roc},
using the predictive confidence from PCNet for computing the set representation  gives an improvement 
for all metrics by about 2-8\% over the raw ResNet50 and SENet50 on 1:1 mixed verification.
The performance improvements are most substantial at low FARs,
this is as expected due to the fact that the main issue of average feature aggregation (as used in previous results) 
is that the set-based representation can be distracted by images of low quality, leading to  false matchings. 
Consequently, the most dramatic improvement is from highlighting the discriminative faces and suppressing the ones with inadequate information.

\input{tables/IJBC_11_template.tex}

\subsection{Visualization.} 
\begin{figure*}[!htb]
\footnotesize
\begin{center}
\includegraphics[width=\textwidth]{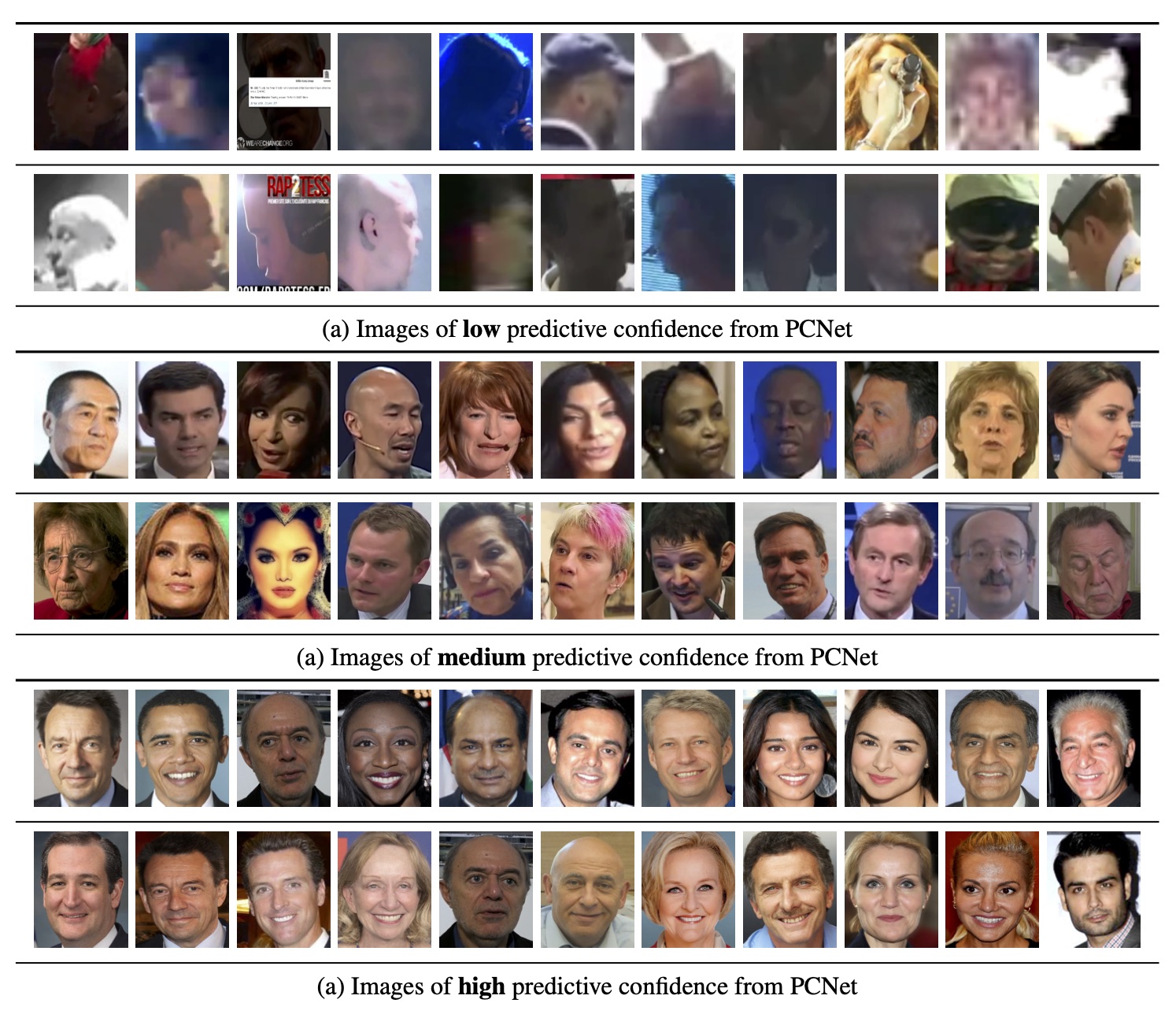} \\
\caption[]{After ranking all images of the IJB-C datasets, 
we split the ranking into three different ranges, and randomly sample images from the corresponding range.}
\vspace{-20pt}
\label{fig:vis}
\end{center}
\end{figure*}

In Figure~\ref{fig:vis}, 
we show the sorted images in ascending order based on the predictive confidences inferred from PCNet.
As expected, the low confidence scores for aberrant images are highly correlated with human expectation,
\ie~blurry, nonface, extreme poses.
Note, the images of medium and high quality are not so well separated, 
though high quality ones are often near frontal. 

%% file: tables/IJBC_11_covariate.tex
\newcommand{\tabincell}[1]{\begin{tabular}{@{}#1@{}}#1\end{tabular}}
\begin{table}[!htb]
\small
\begin{center}{\scalebox{0.8}{
\begin{tabular}{c|c|c|c|c|c|c|c|c|c|c|c}
\hline
 & & 
\multicolumn{5}{c|}{TAR@FAR=1E-6}  & \multicolumn{5}{c}{TAR@FAR=1E-5} \\ 

 Verification Arch. & Predictive Conf. & r=$0.0$ & r=$0.1$ & r=$0.2$ & r=$0.3$ & r=$0.4$ & r=$0.0$ & r=$0.1$ & r=$0.2$ & r=$0.3$ & r=$0.4$ \\ 
\hline
ResNet-50 & MNet~\cite{Xie18b} &  $0.202$ & $0.392$ & $0.446$ & $0.488$ & $0.512$ & $0.433$ & $0.578$ & $0.616$ & $0.638$ & $0.657$ \\ 
ResNet-50 & QNet~\cite{Hernandez-Ortega19} &  $0.202$ & $0.447$ & $0.502$ & $0.544$ & $0.577$ & $0.433$ & $0.613$ & $0.653$ & $0.693$ & $0.728$ \\ 
ResNet-50 &\textbf{PCNet~(Ours)} & 0.202 & \textbf{0.510} & \textbf{0.536} & \textbf{0.572} & \textbf{0.617} & {0.433} & \textbf{0.641} & \textbf{0.673} & \textbf{0.718} & \textbf{0.769} \\ \hline

SENet-50 & MNet~\cite{Xie18b} &  $0.317$ & $0.399$ & $0.423$ & $0.455$ & $0.487$ & $0.528$ & $0.595$ & $0.621$ & $0.648$ & $0.670$ \\ 
SENet-50 & QNet~\cite{Hernandez-Ortega19} &  $0.317$ & $0.453$ & $0.481$ & $0.517$ & $0.539$ & $0.528$ & $0.629$ & $0.660$ & $0.695$ & $0.724$ \\ 
SENet-50 & \textbf{PCNet~(Ours)} &  {0.317} & \textbf{0.478} & \textbf{0.508} & \textbf{0.538} & \textbf{0.578} & {0.528} & \textbf{0.644} & \textbf{0.674} & \textbf{0.715} & \textbf{0.761} \\ \hline


ResNet-101 & MNet~\cite{Xie18b} &  $0.095$ & $0.248$ & $0.366$  & $0.387$ & $0.440$ & $0.249$ & $0.548$ & $0.622$ & $0.659$ & $0.685$ \\ 
ResNet-101 & QNet~\cite{Hernandez-Ortega19} &  $0.095$ & $0.280$ & $0.352$  & $0.409$ & $0.469$ & $0.249$ & $0.592$ & $0.653$ & $0.700$ & $0.738$ \\ 
 ResNet-101 & \textbf{PCNet~(Ours)} & {0.095} & \textbf{0.465} & \textbf{0.490}  & \textbf{0.530} & \textbf{0.565} & {0.249} & \textbf{0.671} & \textbf{0.707} & \textbf{0.752} & \textbf{0.797} \\ \hline

\end{tabular}}}
\end{center}
\caption{Error vs rejection on IJB-C 1:1 covariate verification. 
By only rejecting a small proportion of the low quality image pairs~($r \in [0.0, 0.4]$),
significant performance boost can be observed for all different architectures.}
\label{tab:11cov}
\vspace{-5pt}
\end{table}

%% file: tables/IJBC_11_template.tex
\begin{table}[h]
\small
\begin{center}{\scalebox{0.8}{
\begin{tabular}{l|c|c|c|c|c|c|c}
\hline
  & & & \multicolumn{5}{c}{1:1 Verification TAR} \\
 & Architecture  &Predictive Conf. & FAR=1E-6 & FAR=1E-5 & FAR=1E-4 & FAR=1E-3 & FAR=1E-2 \\
\hline
Cao {\it et al.}~\cite{Cao18}
& ResNet50& - & $0.610$ & $0.742$ & $0.842$ & $0.916$ &  0.958 \\
& ResNet50& QNet~\cite{Hernandez-Ortega19} & $0.641$ & $0.762$ & $0.860$ & $0.929$ & 0.969 \\
&ResNet50& MNet~\cite{Xie18b} & 0.664 & 0.770 & 0.864 & 0.930  & 0.969\\
&ResNet50& \textbf{PCNet~(Ours)} & \textbf{0.693} & \textbf{0.803} & \textbf{0.885} & \textbf{0.944}  & \textbf{0.970} \\

\hline
Cao {\it et al.}~\cite{Cao18}
& SENet50&  & $0.617$ & $0.753$ & $0.852$ & $0.927$ & 0.971 \\
& SENet50& QNet~\cite{Hernandez-Ortega19}  & $0.643$ & $0.768$ & $0.861$ & $0.931$  & 0.972 \\
& SENet50& MNet~\cite{Xie18b} &  $0.649$ & $0.775$ & $0.867$ & $0.932$  & 0.973\\
& SENet50& \textbf{PCNet~(Ours)} &  \textbf{0.695} & \textbf{0.800} & \textbf{0.890} & \textbf{0.948}  & \textbf{0.974} \\
\hline
\end{tabular}}}
\end{center}
\caption[caption]{
\footnotesize{Evaluation on 1:1 verification protocol on IJB-C dataset. Higher is better.
The numbers with MNet~\cite{Xie18b} are based on our re-implementations,
and QNet is from the official implementation and model~\cite{qnet}.
}}
\label{tab:ijbc-evaluation-table2}
\vspace{-30pt}
\end{table}


%% file: text/conclusions.tex
To summarize, in this paper, 
we propose a novel training scheme for learning predictive confidence, 
with the goal of reducing the proportion of errors caused by images with insufficient information, 
\eg~poor visual quality or profile faces, non-face images.
While evaluating on the challenging JANUS IJB-C Benchmarks,
we demonstrate three use cases:
 (i) PCNets can be used to significantly improve 1:1 image-based verification error rates, 
of automatic face recognition systems by rejecting low-quality face images,
(ii) it can be used for quality score based fusion where a weighted average is used to compute set representation, 
(iii) it can also be used as a quality measure for selecting good (unblurred, good lighting, more frontal) faces from a collection, 
\eg~for automatic enrollment or display. 
Although we have presented the predicitive confidence as essential for face verification, 
the idea of learning a confidence from true matches is more generally applicable. 
For example, 
a predictive confidence could be learnt from a set of ground truth matches between images, 
and then used to predict a confidence for correspondences between images or video frames.

%% file: tables/architecture.tex
\begin{table}[!htb]
\scriptsize
\renewcommand\arraystretch{1.4}
\begin{center}{\scalebox{0.85}{
\begin{tabular}{c|p{3.5cm}<{\centering}|p{3.5cm}<{\centering}}
\hline
Architecture 
& Input Image ($N \times 224 \times 224 \times 3$) 
& Output Size \\
\hline
\multirow{17}{*}{\rotatebox[origin=c]{90}{Predictive Confidence Network}}
& \multicolumn{1}{c|}{conv, $7\times7$, $64$, stride $2$} 
& $112 \times 112 \times 64$  \\
\cline{2-3}

& \multicolumn{1}{c|}{max\;pool, $3\times3$, stride $2$}
& $56 \times 56 \times 64$  \\
\cline{2-3}

& \multicolumn{1}{c|}
{$\begin{bmatrix} {\rm conv}, 3\times 3, 64 \\ {\rm conv}, 3\times 3, 64  \end{bmatrix} \times 2$}
&$ 56 \times 56 \times 64$  \\
\cline{2-3}

& \multicolumn{1}{c|}
{$\begin{bmatrix} {\rm conv}, 3\times 3, 128 \\ {\rm conv}, 3\times 3, 128  \end{bmatrix} \times 2$}
&$ 28 \times 28 \times 128$    \\
\cline{2-3}

& \multicolumn{1}{c|}
{$\begin{bmatrix} {\rm conv}, 3\times 3, 256 \\ {\rm conv}, 3\times 3, 256  \end{bmatrix} \times 2$}
&$ 14 \times 14 \times 256$    \\
\cline{2-3}

& \multicolumn{1}{c|}
{$\begin{bmatrix} {\rm conv}, 3\times 3, 512 \\ {\rm conv}, 3\times 3, 512 \end{bmatrix} \times 2$}
&$ 7 \times 7 \times 512$    \\
\cline{2-3}

& Global Average Pooling
&$ 1 \times 1 \times 512$    \\
\cline{2-3}

& FC, $1 \times 1$, $128$
&$ 1 \times 1 \times 128$    \\
\cline{2-3}

& Predictive Confidence, $1 \times 1$, $1$
&$ 1 \times 1 \times 1$    \\
\hline

\end{tabular}}}
\end{center}
\caption{Architecture of the proposed Predictive Confidence Network~(PCNet).}
\label{arch}
\vspace{-0.3cm}
\end{table}